\documentclass[letterpaper, 10 pt, conference]{ieeeconf}
\IEEEoverridecommandlockouts

\usepackage{amsmath}
\usepackage{amssymb}
\usepackage{amsfonts}
\usepackage{bm}
\usepackage{cite}
\usepackage{graphicx}
\usepackage{algorithm}
\usepackage{algpseudocode}
\usepackage[colorlinks=true,allcolors=blue]{hyperref}
\usepackage{booktabs}

\newtheorem{proposition}{Proposition}

\newcommand{\reals}{\mathbb{R}}

\newcommand{\tf}{t_f}

\newcommand{\Ps}{P}

\DeclareMathOperator*{\argmin}{arg\,min}

\title{\LARGE \bf Time-Optimal Ergodic Retrieval}

\author{Christian Sulana$^{1}$, Christian Hughes$^{2}$, and Ian Abraham$^{3}$
\thanks{$^{1}$Christian Sulana is with Saddleback College, Mission Viejo, California, USA
        {\tt \small csulana@hotmail.com}}
\thanks{$^{2}$Christian Hughes is with the Department of Mechanical Engineering,
        Yale University, New Haven, CT 06520, USA
        {\tt \small christian.hughes@yale.edu}}%
\thanks{$^{3}$Ian Abraham is with the School of Electrical and Computer Engineering,            The University of Sydney,
        Sydney, Australia
        {\tt \small ian.abraham@sydney.edu.au}}}%

\begin{document}
\maketitle
\thispagestyle{empty}
\pagestyle{empty}

\let\thefootnote\relax\footnotetext{We acknowledge the use of Claude to assist with grammar/correctness checks and code troubleshooting.}

\begin{abstract}
Retrieval tasks are a form of search task where a robot must explore, find an object, and return with it as quickly as possible. 
Recent work in ergodic control has demonstrated time-optimal path planning for search tasks where visitation is proportionally matched, within some tolerance, to a given target distribution. 
However, these planners minimize the duration of a full deterministic search, while retrieval missions end their search at an unknown point in the trajectory and must return to a given base. 
In this work, we propose a new formulation of the time-optimal ergodic search problem that models detection as a Poisson process and minimizes total mission time by including the return phase in the objective.
Experimental results show that time-optimal ergodic retrieval (TOER) achieves up to $19\%$ shorter mission durations relative to ergodic baselines and that this advantage grows as the return phase becomes more costly. 
Simulation and hardware implementations validate TOER's compatibility with environments of non-convex geometry as well as both 2D and 3D search domains. 
\end{abstract}

\section{Introduction}

Retrieval tasks are common in search-and-rescue, logistics, and inspection tasks, where a robot must locate an object and bring it to a given destination \cite{cao2023,englot2014,ivic2023}.
Such missions do not end when the object is found, but rather when it is delivered back to base. 
Because detection is uncertain, the point of transition from search to return is a random variable unknown at the time of planning. 
A planner that optimizes only up to the moment of detection therefore optimizes a fraction of the total mission which is smallest when the object is hardest to find.
In this work, we address the question: \textbf{How should a robot order its search when the cost of completing the mission depends on where the search ends?}

\begin{figure} [t]
    \centering
    \includegraphics[width=\linewidth]{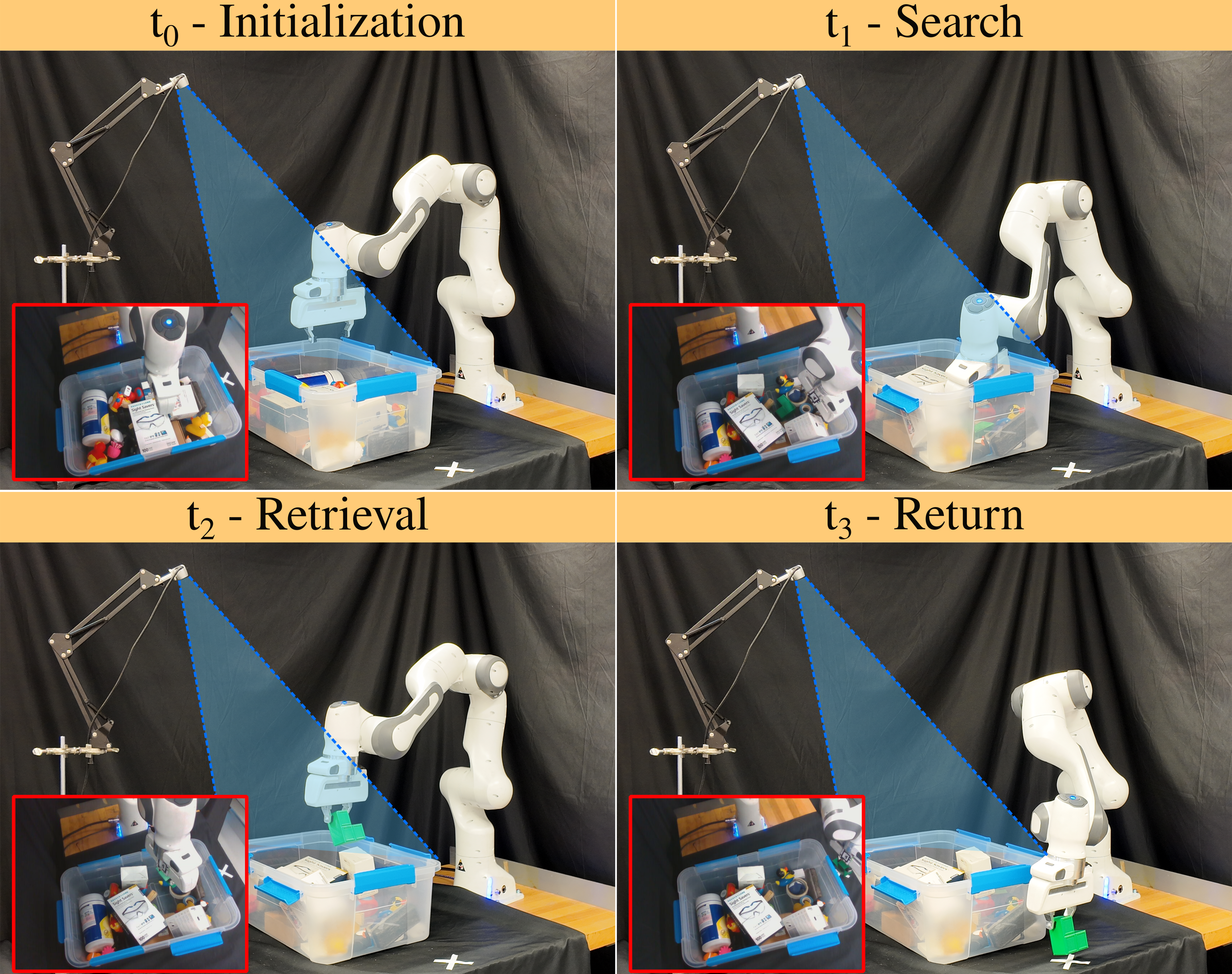}
    \vspace{-16pt}
    \caption{\textbf{Time-Optimal Ergodic Retrieval.} A Franka arm executes a full retrieval mission of a green block in an occluded 3D search domain. 
    From its start pose ($t_0$), the arm sweeps the bin along an ergodic trajectory ($t_1$) until the overhead camera detects the target. 
    Once detected, the arm abandons the remaining plan, grasps the object ($t_2$) and delivers it to the base ($t_3$). 
    We propose time-optimal ergodic retrieval (TOER), which includes the expected return cost explicitly in the objective such that searches prioritize likely detection sites that are closer to the base.}
    \label{fig:abstract}
\end{figure}

Ergodic search methods plan single-shot trajectories whose time-averaged visitation is proportional to a given target distribution on the domain \cite{mathew2011metrics, miller2015ergodic}. 
Kernel-based formulations have extended these guarantees to non-Euclidean domains via the Maximum Mean Discrepancy metric \cite{hughes2024ergodic}, while time-optimal ergodic search methods (TOES) invert the classical objective to minimize terminal time subject to an ergodic inequality constraint to produce the fastest trajectory that still covers the domain to a chosen tolerance \cite{dong2023}.

However, these formulations do not currently represent that the search outcome is stochastic. 
The optimized horizon is the duration of the full planned search, while retrieval missions end at a random point in the path where the target is detected. 
Since the robot rarely executes the entire plan, minimizing the terminal search time generally optimizes an incorrect duration. 
In addition, while current time-optimal formulations define a terminal boundary constraint to force the trajectory to end at a base, this hard constraint fixes the final pose even when detection occurs at an earlier segment. 

In this paper, we propose time-optimal ergodic retrieval (TOER), which minimizes the expected duration of the full search-and-return mission subject to an ergodic constraint. 
We model detection along a trajectory as an inhomogeneous Poisson process whose rate is set by the sensor's response to a given target distribution such that the probability that the robot is still searching decays with accumulated detection pressure. 
The objective sums the expected search time with an expected return time charged from the node at which detection occurs. 
We retain ergodicity as a constraint to preserve the coverage guarantees of time-optimal ergodic search and  eliminate the degenerate solution of immediately abandoning the search. 

We show that the return may be modeled by any lower bound on the return cost, including a straight-line estimate that ignores obstacles, and that the resulting plan carries a computable certificate of $\varepsilon$-optimality. 
Constraining the survival probability at the horizon certifies that the target is detected with at least the given probability. 
We prove that ergodic coverage guarantees transfer unchanged, and that the value of return-awareness scales predictably with the difficulty of the return toward a computable ceiling fixed by domain geometry, coverage tolerance, and the target distribution. 

In summary, the contributions of this work are:
\begin{itemize}
    \item A novel formulation of ergodic retrieval that minimizes the expected duration of a search-and-return mission by charging the return from an unknown detection point;
    \item Derived guarantees that any admissible lower bound on the return yields a certifiably $\varepsilon$-optimal plan, that ergodic coverage guarantees hold in this formulation, and that return-awareness' advantage grows with return discrepancy toward a fixed ceiling;
    \item Validation in 2D and 3D domains in simulation and on hardware, including isolation of the return term against a search-only reference. 
\end{itemize}

\section{Related Work}

\subsection{Search Under Detection Uncertainty}

Search under uncertainty has classically been posed as an allocation of limited effort to maximize the likelihood of detection. 
Search theory models a sensor's chance of finding a target through an exponential detection law whose exponent accumulates with the effort applied in a region \cite{koopman1980search}.
Recent work uses a Poisson structure to tie search effort to the probability that an object remains undetected \cite{jovan2023efficiently}. 
This approach gives a principled notion of how detection likelihood accumulates during search, but is posed over the allocation of sensing effort instead of a trajectory.

Information-driven exploration instead scores uncertainty directly and drives a robot to reduce it \cite{bourgault2002}. 
Existing work leverages uncertainty metrics such as expected information gain \cite{schlotfeldt2019}, the posterior variance of a Gaussian process \cite{jadidi2014}, or learned predictions of unobserved regions from sparse samples \cite{caley2020}. 
These scores are used greedily \cite{jadidi2014} or over a short lookahead planned via Bayesian optimization \cite{marchant2014}. 
However, since the decision horizon is often short, the resulting paths are locally efficient but make no guarantees regarding the coverage that the trajectory induces over the domain. 

\subsection{Ergodic Search Methods}
Ergodic search methods resolve the short-horizon problem of greedy methods by constructing trajectories whose time-averaged visitation proportionally matches a given target distribution \cite{mathew2011metrics, miller2015ergodic}. 
Traditional ergodic metrics are defined over Fourier spectral modes, which fix the approach to a rectangular coordinate structure. 
However, recent work has shown that kernel-based formulations such as Maximum Mean Discrepancy (MMD) extend ergodic coverage to domains of arbitrary geometry \cite{hughes2024ergodic} and scale \cite{lahrach2025}, as well as flow-evolved search spaces \cite{hughes2026asymptoticallyoptimalergodiccoverage}. 

Time-optimal ergodic search inverts the classical ergodic problem by minimizing overall search time $t_f$ subject to an ergodic constraint $\mathcal{E}\leq\gamma$ \cite{dong2023}. 
The resulting trajectory is the fastest path that still reaches a chosen level of coverage, as characterized through Pontryagin's conditions of optimality. 
However, these formulations fix the final trajectory point via a terminal constraint and treat the search horizon as fixed once planned, which is incompatible with retrieval mission priorities. 
Because object detection can happen at any segment of the search trajectory, minimizing $t_f$ optimizes a duration that is rarely realized, and when the search ends early, the terminal constraint is fixed to a trajectory point that the robot never reaches. 


\subsection{Return-Aware Planning}
Planning that accounts for a return trip is well-studied when the destination is known in advance. 
Orienteering and prize-collecting formulations select which locations to visit under a budget that must accommodate the trip back to base \cite{vansteenwegen2011}, and coverage planners for energy-limited platforms impose a return-to-base condition on the terminal state \cite{franco2015}. 
In mobile manipulation, search-and-return pipelines typically decompose the task into separate search and transport phases that execute sequentially once the object has been located \cite{danielczuk2019}. 
However, each of these approaches assumes that the return originates from a location that is either fixed or chosen by the planner. 
In retrieval under detection uncertainty, the return instead originates from wherever detection occurs, which is a random variable induced by the search trajectory. 
In this work, we compute a mission's return as the expected return cost from each trajectory point to account for the unknown target location. 


\section{Preliminaries}

\subsection{Poisson Process}
\label{sec:poisson-process}
    A Poisson process is a memoryless stochastic model for counting discrete random events that occur independently in continuous time at a given rate. 
    Let $N(t)$ be the number of events that occur in the interval $[0,t]$. For a varying rate, $N(t)$ follows a Poisson distribution with mean $\Lambda(t)$:
    \begin{equation}
        \Pr\big(N(t) = n\big) = \frac{\Lambda(t)^{n}\, e^{-\Lambda(t)}}{n!},
        \label{eq:poisson-pmf}
    \end{equation}
    where $\Lambda(t)$ is the accumulated probability of an event's occurrence. The probability that no events happen by time $t$ is $\Pr\big(N(t)=0\big)=e^{-\Lambda(t)}$, which decays exponentially as $\Lambda(t)$ grows.

\subsection{Ergodicity}
\label{sec:ergodicity}
    In robotic search problems, ergodicity is often used to describe the extent to which a robot's time spent in the search domain proportionally represents the prior utility of visiting each region~\cite{mathew2011metrics}. In exploration cases, an ergodic trajectory samples or spends time in each region in proportion to a target distribution $\mathcal{I}$.
    Let the time-averaged trajectory distribution on $\Omega$ be of the discrete form:
    \begin{equation}
        \rho_{\bar{x},t_f}(\omega)=\frac{1}{t_f}\sum_{t=0}^{t_f-1}\delta[\omega-g(\bar{x}(t))],
    \label{eq:ergodicity}
    \end{equation}
    where $g$ defines robot's position in coverage space and $\delta$ is the Dirac-delta function. 
    
    In practice the horizon is finite, so exact ergodicity is unattainable; an ergodic metric is instead minimized so that $\rho_{\bar{x},t_f}$ converges to $\mathcal{I}$ closely over a finite horizon $t_f$.

\subsection{Ergodic Control}
\label{sec:ergodic-control}
    Ergodicity is enforced through a metric that scores the distance between $\rho_{\bar{x}}$ and $\mathcal{I}$ under a functional transform $F$:
    \begin{equation}
        \mathcal{E}(\bar{x}, \mathcal{I}) = \big\lVert F[\rho_{\bar{x}}] - F[\mathcal{I}] \big\rVert_F^2,
    \label{eq:ergodic-metric}
    \end{equation}
    with $\lVert\cdot\rVert_F$ as the associated norm~\cite{hughes2026asymptoticallyoptimalergodiccoverage}. Classically, $F$ is the Fourier transform~\cite{mathew2011metrics} or a kernel-based sample map~\cite{hughes2024ergodic}. 
    For any ergodic metric, $\bar{x}$ is ergodic with respect to $\mathcal{I}$ if and only if $\mathcal{E} \to 0$ as $t_f \to \infty$.

    In practice, ergodic planners generate trajectories by solving the optimization problem:
    \begin{equation}
        \begin{aligned}
            \min_{\bar{x},u}\quad & \mathcal{E}(\bar{x}) \\
            \text{s.t.}\quad & h_1(\bar{x}_t, u_t) = 0\ \forall\, t \in [0, t_f - 1] \\
            & h_2(\bar{x}_t, u_t) \le 0\ \forall\, t \in [0, t_f - 1],
        \end{aligned}
    \end{equation}
     where $\bar{x}$ and $u$ are the state and control trajectories and $h_1,\ h_2$ are the constraints.
    Recent time-optimal formulations instead represent the ergodic control problem as the following:
    \begin{equation}
        \begin{aligned}
            \min_{x(t),u(t)}\quad & t_f \\
            \text{s.t.}\quad & \mathcal{E}\leq\gamma, \quad t_f>0, \\
            & h_1(x,u) = 0, \quad h_2(x,u) \le 0.
        \end{aligned}
    \end{equation}
    Such formulations maintain ergodic convergence guarantees within a threshold $\gamma$ while generating trajectories that are time-optimal with respect to the search process~\cite{dong2023}.

    This work's method retains the structure of a minimized time cost subject to $\mathcal{E}\leq\gamma$ but replaces the deterministic horizon $t_f$ with the expected duration of the search (whose stopping time is unknown) and adding the cost of returning from wherever that stop occurs. 
    
\subsection{Maximum Mean Discrepancy}
\label{sec:mmd}
    Maximum Mean Discrepancy (MMD) is a two-sample statistical metric that quantifies the distances between two probability distributions by comparing their kernel mean embeddings in a Reproducing Kernel Hilbert Space (RKHS).
    Let $k:\Omega \times\Omega \rightarrow \reals^{\geq0}$ be a positive kernel with associated RKHS $\mathcal{H}$, so that a distribution $p$ embeds as $\mu_p=\mathbb{E}_{\omega\sim p}[k(\omega,\cdot)]$ and $\text{MMD}_k^2(p,q)=\lVert\mu_p-\mu_q\rVert_\mathcal{H}^2$. 
    The kernel is called \emph{characteristic}~\cite{sriperumbudur2011} when this mapping is injective, which allows $p$ to be a unique representation in $\mathcal{H}$.
    For finite samples $\mathbf{x} = \{x_i\}_{i=1}^N$ and $\mathbf{y}=\{y_j\}_{j=1}^M$ with $x_i\sim p$ and $y_j\sim q$, the empirical approximation of the MMD can be calculated as:
    \begin{equation}
        \begin{split}
        \mathrm{MMD}^2_k(\mathbf{x},\mathbf{y})
            ={}  \frac{1}{N^2}\sum_{i=1}^{N}\sum_{i'=1}^{N} k(x_i,x_{i'}) \\
             - \frac{2}{NM}\sum_{i=1}^{N}\sum_{j=1}^{M} k(x_i,y_j)
            + \frac{1}{M^2}\sum_{j=1}^{M}\sum_{j'=1}^{M} k(y_j,y_{j'}).
        \end{split}
    \label{eq:mmd-empirical}
    \end{equation}
    The advantage of the MMD ergodic metric $\left(\text{MMD}^2(\bar{x}, \mathcal{I})=\mathcal{E}(\bar{x}, \mathcal{I})\right)$ is its generalizability to domains of arbitrary geometry, which extends ergodic coverage to any traversable domain where spatial samples can be collected.     
\section{Time-Optimal Ergodic Retrieval}
\subsection{Problem Setup}
    Consider a robot searching a known, static, bounded domain $\Omega \subset \mathbb{R}^n$ for an object at an unknown location. Prior assumptions of the object's location, when available, are encoded as a discrete distribution over $M$ candidate locations $\{y_j\}_{j=1}^{M}$ with weights $w_j \ge 0$ and $\sum_{j=1}^{M} w_j = 1$, approximating the underlying target distribution $\mathcal{I}$. The robot travels a piecewise-linear trajectory $\bar{x} = \{x_0, x_1, \dots, x_N\}$, traversing segment $i$ from $x_{i-1}$ to $x_i$ in time $\{\tau_i\}_{i=1}^{N}$ at speeds up to the robot's search actuation limit $v_s$. The robot may start anywhere, and upon finding the object, must deliver it to a fixed base $x_b \in \Omega$.
    
\subsection{Detection Likelihood}
\label{sec:lambda}
    From the Poisson detection model of Section~\ref{sec:poisson-process}, we construct the detection likelihood that drives the objective. When the robot is at $x_i$, its sensor detects an object located at $y_j$ at an instantaneous rate via a sensor modeled in this work as the Gaussian kernel:
    \begin{equation}
        s(x_i, y_j) = s_0 \exp\!\left( -\frac{\lVert c(x_i) - y_j \rVert^2}{h_{\text{sensor}}} \right),
    \label{eq:sensor}
    \end{equation}
    where $s_0$ is the peak detection rate, $h_{\text{sensor}}$ is the sensor's spatial bandwidth, and $c:\mathcal{X}\to\Omega$ maps the robot state to the sensed point in the search domain. 
    The rate reaches $s_0$ when the sensor footprint $c(x_i)$ coincides with the object at $y_j$ and falls off over a length scale $\sqrt{h_{\text{sensor}}}$ with distance. Accumulating this rate over the time spent up through segment $i$ gives the detection pressure at $y_j$ (the discrete form of $\Lambda(t,y)$ in Prop.~\ref{prop:certificate}):
    \begin{equation}
        \Lambda_{i,j} = \sum_{i' \le i} s(x_{i'}, y_j)\,\tau_{i'},
    \label{eq:pressure}
    \end{equation}
    i.e.\ the sensor rate weighted by dwell time and summed along the path, with $s(x_{i'},y_j)$ held constant over each segment. By the Poisson property, the likelihood of missing an object at $y_j$ after accumulating $\Lambda_{i,j}$ is $e^{-\Lambda_{i,j}}$. Marginalizing over the unknown location under the prior gives the survival, or the likelihood the robot is still searching after segment $i$ (the discrete $\Ps$ of Prop.~\ref{prop:certificate}),
    \begin{equation}
        P_i = \sum_{j=1}^{M} w_j\, e^{-\Lambda_{i,j}}.
    \label{eq:survival}
    \end{equation}
    Before any segment is traversed, no pressure has accumulated ($\Lambda_{0,j}=0$), so $P_0=\sum_{j=1}^Mw_j=1$: the robot is guaranteed to continue searching. The likelihood the object is first found during segment $i$ is the drop in survival:
    \begin{equation}
        \pi_i = P_{i-1} - P_i,
        \qquad \sum_{i=1}^{N}\pi_i + P_N = 1 ,
    \label{eq:find}
    \end{equation}
    where $P_N$, the survival after the final segment, is the probability the object is never found.
    These weights enter the objective terms in Section~\ref{sec:objective}.
    
\subsection{Return Cost}
    Unlike search-only planning, our objective must account for delivering the object to $x_b$. 
    Detection may occur on any segment, and the moment it does, the robot abandons the remaining search and returns from that node, meaning the return is enforced from a different node depending on where detection occurs. 
    If no detection occurs, the robot completes its full search and returns from its final node $x_N$.
    In either case, the return cost is the time to travel from the concluding point $x(t)$ back to $x_b$.
    A Euclidean-distance heuristic for this return can be written as:
    \begin{equation}
        R(x(t)) = \frac{\lVert x(t) - x_b\rVert}{v_r}
    \label{eq:return}
    \end{equation}
    for return speed $v_r>0$.
    The Euclidean distance in~\eqref{eq:return} is a placeholder and may be replaced by an obstacle-avoidance path length $d_\Omega(x(t), x_b)$ that routes around obstacles, without changing any other term. The straight-line form is moreover an admissible surrogate for such a path since $\lVert x(t) - x_b\rVert \le d_\Omega(x(t), x_b)$ and it therefore never overstates the true return and preserves the plan's $\varepsilon$-optimality (Prop.~\ref{prop:bound}).

    In scenarios where the return is more difficult, the benefit of return-awareness grows, as represented in Prop. \ref{prop:loading}. 
    For simplicity, this property is represented by artificially decreasing the return velocity $v_r$ with respect to the search velocity $v_s$, though a similar proposition can be framed from the expected return and search times, respectively. 
    Writing $J_s(\bar{x})$ for the expected search time and $\bar{d}(\bar{x})=\mathbb{E}_{y\sim\mathcal{I}}[d(y\mid\bar{x})]$ for the expected return distance from the point at which search ends (both derived in Sec. \ref{sec:objective}), the expected mission time at discrepancy $\kappa=v_s/v_r$ is $J_\kappa(\bar{x})=J_s(\bar{x})+\frac{\kappa}{v_s}\bar{d}(\bar{x})$.

\begin{proposition} [Admissible Return Bound]
\label{prop:bound}
    Call a surrogate return $\underline{d}$ admissible if $0\leq\underline{d}(\bar{x})\leq\bar{d}(\bar{x})$ for every feasible $\bar{x}$, and let $\underline{J}_\kappa(\bar{x}):=J_s(\bar{x})+\frac{\kappa}{v_s}\underline{d}(\bar{x})$ be the surrogate objective the planner actually minimizes. 
    Write $\bar{x}^\dagger$ for a minimizer of $\underline{J}_\kappa$, $J_\kappa^*:=\min_{\bar{x}}J_\kappa(\bar{x})$, and:
    \begin{equation}
    \label{eq:slack}
        \varepsilon:=\frac{\kappa}{v_s}\left(\bar{d}(\bar{x}^\dagger)-\underline{d}(\bar{x}^\dagger)\right) \geq 0.
    \end{equation}
    Then:
    \begin{enumerate}
        \item $\underline{J}_\kappa(\bar{x}) \leq J_\kappa(\bar{x})$ for every feasible $\bar{x}$ and $\min_{\bar{x}} \underline{J}_\kappa (\bar{x}) \leq J_\kappa^*$;
        \item the returned plan is $\varepsilon\text{-optimal}$, $J_\kappa(\bar{x}^\dagger) \leq J_\kappa^* + \varepsilon$, and $\varepsilon$ is computable once $\bar{x}^\dagger$ is known;
        \item the geodesic distance in $\Omega$ dominates the straight-line distance, $\|x-x_b\| \leq d_\Omega(x, x_b) \forall x\in\Omega$, so both induce admissible surrogates. The geodesic surrogate certifies $\varepsilon=0$, while the obstacle-blind surrogate certifies only $\varepsilon\geq0$, with equality only if every geodesic to $x_b$ is a straight segment. 
    \end{enumerate}
\end{proposition}

\begin{proof}
    Admissibility gives $\frac{\kappa}{v_s}\underline{d}(\bar{x}) \leq \frac{\kappa}{v_s}\bar{d}(\bar{x})$ since $\kappa, v_s>0$, and the search terms agree, so $\underline{J}_\kappa \leq J_\kappa$ pointwise. 
    Minimizing both sides preserves the inequality. 
    By \eqref{eq:slack}, $J_\kappa(\bar{x}^\dagger)=\underline{J}_\kappa(\bar{x}^\dagger)+\varepsilon$. 
    Optimality of $\bar{x}^\dagger$ for $\underline{J}_\kappa$ and then item (1) gives $\underline{J}_\kappa(\bar{x}^\dagger)\leq\underline{J}_\kappa(\bar{x}^*_\kappa)\leq J_\kappa(\bar{x}^*_\kappa)=J_\kappa^*$ for any minimizer $\bar{x}^*_\kappa$ of $J_\kappa$. 
    Any path in $\Omega$ from $x$ to $x_b$ is at least as long as the segment joining its endpoints, so $\|x-x_b\|\leq d_\Omega(x, x_b)$, with equality when that segment lies in the free space.
    Since the true return follows the geodesic, the geodesic surrogate has $\underline{d}=\bar{d}$, so \eqref{eq:slack} gives $\varepsilon=0$.
    The straight-line surrogate satisfies $\underline{d}\leq\bar{d}$ and is therefore admissible, and \eqref{eq:slack} gives $\varepsilon\geq0$ strictly wherever a geodesic path to $x_b$ leaves the straight segment. 
\end{proof}

Correctness in Prop. \ref{prop:bound} items (1) and (2) hold for any admissible surrogate regardless of tightness. 
However, item (3) shows that the certified slack grows as the surrogate loosens. 
We do not claim that realized mission time improves monotonically as the surrogate tightens.

\begin{proposition} [Return Discrepancy Profile]
\label{prop:loading}
    Let $v_s$ and $v_r$ be the search and return speeds, respectively, where $\kappa=v_s/v_r$ is the ratio of search to return speed. 
    Let $\bar{x}_0\in\argmin\{J_s(\bar{x}):\bar{x} \text{ feasible for } \eqref{eq:opt}\}$ be the return-blind plan (\eqref{eq:objective} with return removed), and let $J_{\kappa}^0:=J_{\kappa}(\bar{x}_0), J_{\kappa}^*=\min_{\bar{x}} J_{\kappa}(\bar{x})$, and $\bar{d}_{\min}:=\min_{\bar{x}}\bar{d}(\bar{x})$.
    Then:
    \begin{enumerate}
        \item   $J_\kappa^*$ is concave and nondecreasing in $\kappa$;
        \item   the absolute advantage $A(\kappa):=J_\kappa^0-J_\kappa^*$ is nonnegative, nondecreasing, and convex, with $A(0)=0$;
        \item   the relative advantage $R(\kappa):=A(\kappa)/J_\kappa^0$ satisfies $R(0)=0$ and:
        \begin{equation}
        \label{eq:loading}
            R(\kappa)\leq 1-\frac{\bar{d}_{\min}}{\bar{d}(\bar{x}_0)} \forall \kappa, \qquad \lim_{\kappa\to\infty} R(\kappa)=1 - \frac{\bar{d}_{\min}}{\bar{d}(\bar{x}_0)}.
        \end{equation}
    \end{enumerate}
\end{proposition}

\begin{proof}
    For fixed $\bar{x}$, $J_\kappa(\bar{x})$ is affine and nondecreasing in $\kappa$ since $\bar{d}\geq0$. 
    $J_\kappa^*$ is a pointwise minimum of such functions, thus concave and nondecreasing. 
    $\bar{x}_0$ is feasible for TOER from Prop. \ref{prop:ergodic}, so $J_\kappa^* \leq J_\kappa^0$ and $A\geq0$. 
    $A$ is affine minus concave, hence convex, and $A(0)=0$ since at $\kappa=0$ both problems reduce to $\min J_s$. 
    Convexity with $A(0)=0$ and $A\geq0$ gives $A(\kappa_1)\leq(\kappa_1/\kappa_2)A(\kappa_2)\leq A(\kappa_2)$ for $\kappa_1<\kappa_2$, so $A$ is nondecreasing and $R(0)=0$ follows from $A(0)=0$. 
    Let $\bar{x}^*_\kappa$ attain $J_{\kappa}^*$. 
    Optimality of $\bar{x}_0$ for $J_s$ gives $J_s(\bar{x}_0)\leq J_s(\bar{x}^*_\kappa)$, so $A(\kappa)\leq\tfrac{\kappa}{v_s}\!\left(\bar{d}(\bar{x}_0)-\bar{d}_{\min}\right)$, and dividing by $J_\kappa^0\geq\frac{\kappa}{v_s}\bar{d}(\bar{x}_0)$ gives the bound in \eqref{eq:loading}.
    Finally, let $\bar{x}_d$ attain $\bar{d}_{\min}$.
    Optimality of $\bar{x}_\kappa^*$ against $\bar{x}_d$ gives $\frac{\kappa}{v_s}(\bar{d}(\bar{x}_\kappa^*)-\bar{d}_{\min})\leq J_s(\bar{x}_d)-J_s(\bar{x}_0)$, so $\bar{d}(\bar{x}_\kappa^*)\to\bar{d}_{\min}$ and $R(\kappa)\to1-\bar{d}_{\min}/\bar{d}(\bar{x}_0)$.

\end{proof}

Note that the intrinsic behavior of $J_\kappa^*$ is its concavity and is not fixed to a specific functional form.

Prop. \ref{prop:loading} characterizes the advantage over the ideal return-blind planner. 
For an arbitrary feasible reference, the profile remains convex and nonnegative but need not vanish at $\kappa=0$, therefore monotonicity holds only beyond its minimizer and the ceiling \eqref{eq:loading} holds asymptotically. 

\subsection{Objective}
\label{sec:objective}
    The objective is the expected total mission time: the time spent searching plus the time spent returning the object to $x_b$. We build it from the likelihoods of Section~\ref{sec:lambda}.

    Segment $i$ contributes its time $\tau_i$ only if the robot is still searching when it begins $(P_{i-1})$.
    If the object is first found in segment $i$ (probability $\pi_i$), the robot returns from $x_i$ at cost $R(x_i)$. 
    If it is never found (probability $P_N$), the robot returns from its final node $x_N$ with cost $R(x_N)$. 
    As these outcomes partition the possibilities $(\sum_{i=1}^N\pi_i +P_N=1)$, the total mission time is the sum:
      \begin{equation}
        J(\bar{x}) = \sum_{i=1}^{N} \tau_i\, P_{i-1}
        + \sum_{i=1}^{N} \pi_i\, R(x_i)
        + P_N\, R(x_N),
    \label{eq:objective}
    \end{equation}
    where the first term is the expected search time and the remaining two are the expected return time. 

    A detection within segment $i$ is credited to its arrival node $x_i$, so the robot completes the segment and, if successful, returns from $x_i$.
    Since the sensor has range, $x_i$ need not be the object's location.
    For find-and-report missions, or platforms whose collection reach is comparable to the sensor, the return is charged from $x_i$ as in \eqref{eq:objective}, while a limited-reach mechanism adds another term to charge travel from $x_i$ to the object and then to base. 

\begin{proposition} [Detection Certificate]
\label{prop:certificate}
    Let detection be an inhomogeneous Poisson process with rate $s(x,y)\geq0$ for a target at $y\in\Omega$ observed from state $x$, and let:
    \begin{equation}
    \label{eq:pressure_continuous}
        \Lambda(t,y):=\int_0^ts(x(\zeta), y)\,d\zeta
    \end{equation}
    be the accumulated detection pressure along $\bar{x}$. 
    With $\tau$ the acquisition time, the survival probability is $\Ps(t):=\Pr(\tau>t)=\mathbb{E}_{y\sim\mathcal{I}}[e^{-\Lambda(t,y)}]$. 
    Then:
    \begin{enumerate}
        \item $\Ps(0)=1$, $\Ps$ is nonincreasing, and $\Pr(\tau\leq t_f)=1-\Ps(t_f)$;
        \item any plan satisfying $\Ps(t_f)\leq\delta$ for $\delta\in(0,1)$ detects the target with probability at least $1-\delta$;
        \item writing $J_\kappa^*(\delta)$ for the minimum of $J_\kappa$ subject to $\Ps(t_f)\leq\delta$, the constrained feasible sets are nested in $\delta$, so $\delta'\leq\delta$ gives $J_\kappa^*(\delta')\geq J_\kappa^*(\delta)$.
    \end{enumerate}
\end{proposition}

\begin{proof}
    $\Lambda(0,y)=0$ gives $\Ps(0)=\mathbb{E}_{y\sim\mathcal{I}}[1]=1$. Since $s\geq0$, $\Lambda(\cdot,y)$ is nondecreasing, so $e^{-\Lambda(\cdot,y)}$ is nonincreasing for every $y$ and so is its expectation. 
    The identity $\Pr(\tau\leq t_f)=1-\Ps(t_f)$ is the definition of a survival function, and together with the constraint $\Ps(t_f)\leq\delta$ it yields $\Pr(\tau\leq t_f)\geq1-\delta$.
    If $\delta'\leq\delta$ then $\{\bar{x}:\Ps(t_f)\leq\delta'\} \subseteq \{\bar{x}:\Ps(t_f) \leq \delta\}$, and minimizing $J_\kappa$ over a subset cannot yield a smaller value. 
\end{proof}

The certificate is conditional on the target distribution $\mathcal{I}$ and the rate model $s$. 
Sec. \ref{sec:results} reports how it degrades when both are misspecified. 
Item (3) also implies that any $\delta$ below the smallest survival a feasible plan attains is infeasible, which is checked by minimizing $\Ps(t_f)$ alone. 

\subsection{Optimization Problem}
\label{sec:opt}
    We seek the trajectory and segment times that minimize the expected mission time~\eqref{eq:objective} subject to three constraints. Kinematically, each segment must be traversable within its allotted time at speed at most $v_s$, and segment times are nonnegative (shown in \eqref{eq:opt}).

    Ergodically, the trajectory must cover the target $\mathcal{I}$ to tolerance $\gamma$ in the metric of~\eqref{eq:ergodic-metric}, as introduced in Section~\ref{sec:ergodic-control}. The time-optimal ergodic retrieval problem is then
    \begin{equation}
    \begin{aligned}
        \min_{\bar{x},\,\{\tau_i\}} \quad & J(\bar{x}) \\
        \text{s.t.}\quad
        & \lVert x_i - x_{i-1}\rVert \le v_s\,\tau_i, \;\; \tau_i \ge 0, \;\; i = 1,\dots,N,\\
        & \mathcal{E}(\bar{x},\mathcal{I}) \le \gamma .
    \end{aligned}
    \label{eq:opt}
    \end{equation}

    When a detection guarantee is required, the survival constraint $\Ps(t_f)\leq\delta$ of Prop. \ref{prop:certificate} is appended to \eqref{eq:opt}. 
    Minimizing expected mission time alone admits a degenerate solution of giving up immediately and returning without the target. 
    However, this is avoided via the ergodic constraint that demands proportional coverage to tolerance $\gamma$, which allows optimization to change the order in which the domain is covered without sacrificing the extent of the coverage. 

\begin{proposition} [Ergodic Guarantee Maintenance]
\label{prop:ergodic}
    Let the search domain $\Omega$ be compact and let $k:\Omega\times\Omega\to\mathbb{R}$ be a bounded, continuous, characteristic kernel with RKHS $\mathcal{H}$ and mean embedding $\mu_p=\mathbb{E}_{\omega\sim p}[k(\omega,\cdot)]$.
    For a trajectory $\bar{x}$ of duration $t_f$ with state $x(t)\in \Omega$, let $\rho_{\bar{x}}$ be its time-averaged measure and let:
    \begin{equation}
        \mathcal{E}(\bar{x}, \mathcal{I}):=\|\mu_{\rho_{\bar{x}}} - \mu_\mathcal{I} \|^2_\mathcal{H}
    \end{equation}
    be the ergodic metric against the target distribution $\mathcal{I}$. 
    If $\bar{x}$ satisfies the coverage constraint $\mathcal{E}(\bar{x}, \mathcal{I})\leq\gamma$, then:
    \begin{enumerate}
        \item for every $h \in \mathcal{H}$, 
        \begin{equation}
            \left |\frac{1}{t_f} \int_0^{t_f} h(x(t))\,dt - \mathbb{E}_{y\sim\mathcal{I}}[h(y)]\right | \leq \sqrt{\gamma} \|h\|_\mathcal{H};
        \end{equation}
        \item $\gamma\to0$ implies $\rho_{\bar{x}}\rightharpoonup\mathcal{I}$, so $\gamma$ tunes proximity to exact ergodicity;
        \item $\mathcal{E}$ carries no return term, so search-only and return-aware planning share the same feasible set. 
    \end{enumerate}
    Any return-aware plan meeting tolerance therefore inherits (1) and (2), and the search-only optimum remains feasible for the return-aware problem.
\end{proposition}

\begin{proof}
    By the reproducing property and linearity of the integral, $\langle\mu_{\rho_{\bar{x}}}, h\rangle_\mathcal{H} = \frac{1}{t_f} \int_0^{t_f} h(x(t)) dt$ and $\langle\mu_\mathcal{I}, h\rangle_\mathcal{H} = \mathbb{E}_{y\sim\mathcal{I}}[h(y)]$, so their difference is $\langle \mu_{\rho_{\bar{x}}} - \mu_\mathcal{I}, h\rangle_\mathcal{H}$. 
    Cauchy-Schwarz bounds this by $\|\mu_{\rho_{\bar{x}}} - \mu_\mathcal{I}\|_\mathcal{H} \|h\|_\mathcal{H} = \sqrt{\mathcal{E}(\bar{x}, \mathcal{I})} \|h\|_\mathcal{H} \leq \sqrt{\gamma} \|h\|_\mathcal{H}$. 
    A characteristic kernel embeds probability measures injectively, so $\|\mu_{\rho_{\bar{x}}} - \mu_\mathcal{I}\|_\mathcal{H}=0$ only when $\rho_{\bar{x}}=\mathcal{I}$. 
    For a bounded continuous characteristic kernel on compact $\Omega$, this norm metrizes the weak topology on probability measures over $\Omega$ and driving $\gamma$ to zero drives $\rho_{\bar{x}}$ to $\mathcal{I}$ weakly. 
    $\mathcal{E}(\bar{x}, \mathcal{I})$ depends on $\bar{x}$ only through $\rho_{\bar{x}}$ and involves neither the return cost nor $\kappa$. 
    Return-awareness in the objective leaves the constraint set unchanged, which gives the stated feasibility.
\end{proof}


\section{Results}
\label{sec:results}

\begin{figure} [t]
    \centering
    \vspace{14pt}
    \includegraphics[width=\linewidth]{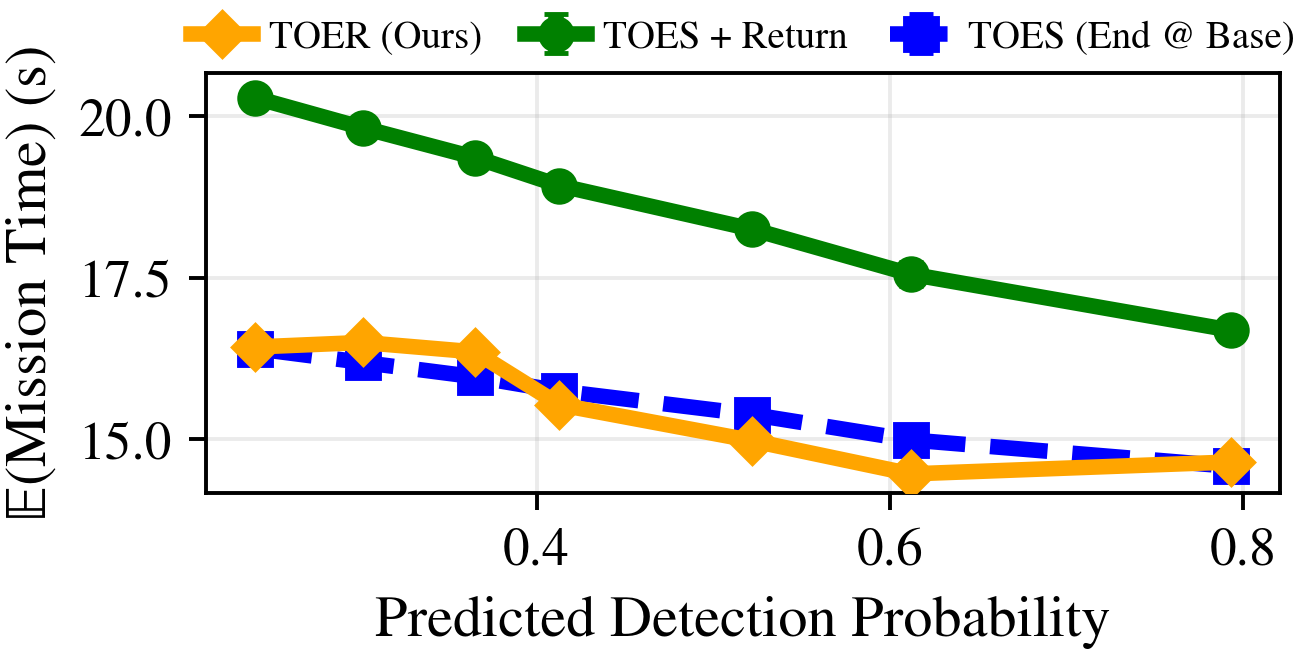}
    \vspace{-20pt}
    \caption{\textbf{Expected Mission Time Comparison.} Single-target search in an open planar field as the sensor's detection gain is swept at fixed $\kappa=6.6$. When detection is unreliable, the target is usually not found, thus mission time is dominated by return, and TOES + Return suffers additional return cost at every unsuccessful episode. TOER holds a $\geq12\%$ advantage over TOES + Return throughout while matching TOES (End @ Base) without imposing a terminal constraint.}
    \label{fig:e1}
\end{figure}


We compare our method against the following baselines: time-optimal ergodic search with a heuristic return [TOES + Return], TOES with a hard terminal constraint at the base [TOES (End @ Base)], Info-Max with a heuristic return, and TOER with the return term removed from its objective [TOER (search-only)]. 
All methods share the same point-mass dynamics with speed limit $v_s=1.4$ m/s, sensor model, solver budget, and discretization of $N=500$ knots over $M=1500$ domain samples, and each is scored with the same geodesic return from its realized collection point. 
Unless otherwise specified, every method is given a uniform target distribution. 
Ergodic methods impose the proportionality constraint with the same Augmented Lagrangian solver using a Gaussian kernel with $h=0.03$ and tolerance $\gamma=2.5$ unless otherwise stated, and obtain matched realized ergodicities within $0.01\%$.
Each run is solved from a shared bank of $10$ initializations and we report the one with the lowest expected mission time. 
Metrics are estimated by Monte-Carlo over 2000 target draws with detection simulated as the true Poisson process, and all intervals are $95\%$ paired bootstrap confidence intervals over the shared target draws. 


\subsection{Return-Awareness Impact}
\label{sub:crossover}

We first compare ergodic methods across detection reliability by sweeping the sensor's detection gain at a fixed return discrepancy $\kappa=6.6$ such that the likelihood of locating the target ranges from $[0.24, 0.79]$ (see Fig. \ref{fig:e1}). 

\textbf{Problem Setup.} A robot is placed in an open planar field with the domain $[-1.4, 1.4]^2$m, starting at the base located at $(-1.28, -1.28)$.
The sensor is modeled as a Gaussian kernel with $h_\text{sensor}=0.048$, and the detection gain is scaled over $[1.2, 12]$ with coverage tolerance $\gamma=2.8$.

As shown in Fig. \ref{fig:e1}, TOER's advantage is largest when detection is least reliable. 
At $P(\text{found})=0.24$, TOER completes the mission 19\% faster than TOES + Return, and TOER holds a $\geq12\%$ advantage over TOES + Return at every detection level measured. 

Return-awareness matters most in deciding where the plan terminates, and when the mission ends without object detection, terminal position dominates the expected mission length. 
TOES + Return finishes in this case $4.8$s from base and rises to $8$s at the highest gain, while TOER ends within $0.1$s of base at all but the highest detection level. 
TOER matches TOES (End @ Base) within $\approx3.5\%$ across the sweep without a terminal constraint, which would otherwise force the search to end at $x_b$ even when detection occurs at segment $i\ll N$.


\subsection{Return Discrepancy Profile}
\label{sub:loading}

\begin{figure} [t]
    \centering
    \vspace{14pt}
    \includegraphics[width=\linewidth]{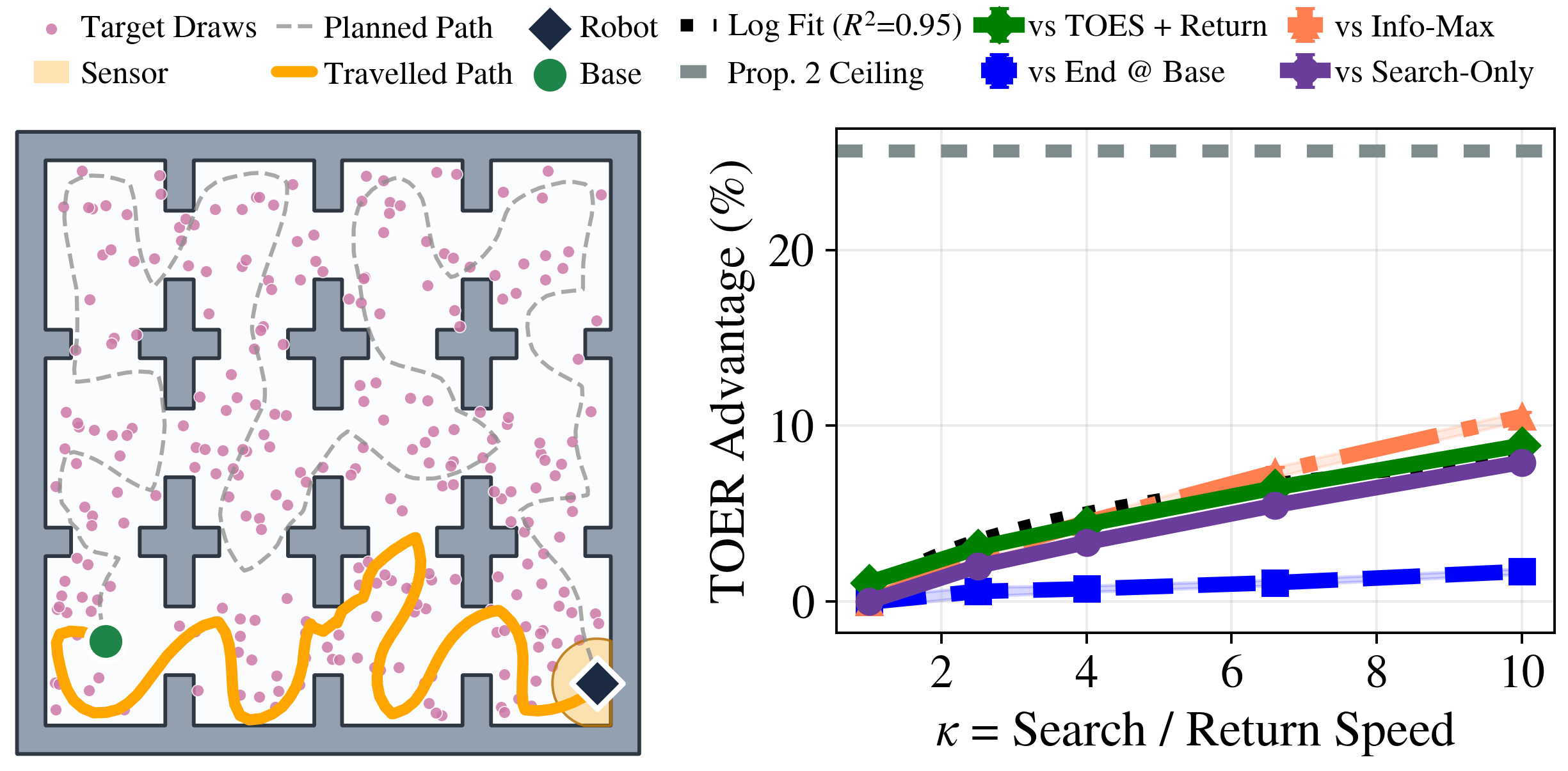}
    \vspace{-18pt}
    \caption{\textbf{Return Discrepancy Profile.} Object search in a planar 12-room lattice. (Left) Search scene with uniform target draws and sensor footprint, occluded by walls. (Right) TOER's paired mission-time advantage over each baseline vs. return discrepancy $\kappa$. Advantage grows logarithmically ($R^2=0.95$) toward the ceiling from Prop. \ref{prop:loading}. The gap to TOER (search-only) isolates the return term, which is null at $\kappa=1$ and $7.9\%$ at $\kappa=10$.}
    \label{fig:e3}
\end{figure}

We next examine the difference in TOER's advantage against baselines as the return discrepancy $\kappa$ is varied from $[1, 10]$. 

\textbf{Problem Setup.} We place a robot with point-mass dynamics in a $2.8\times2.8$m planar domain segmented into 12 rooms. 
Each interior room is open at each cardinal direction so that maneuverability is constrained while path selection stays varied.
The sensor model uses $h_\text{sensor}=0.048$ with a gain of $1.7$ so that detection is unreliable and a mission that ends without finding the object is dominated by the return. 

As shown in Fig. \ref{fig:e3}, TOER's advantage grows with $\kappa$ as described in Prop. \ref{prop:loading}, from $1.1\%\to8.9\%$ over TOES + Return and $0\%\to1.7\%$ over TOES (End @ Base). 
For every $\kappa\geq2.5$, TOER is faster than all four baselines with confidence bands excluding zero. 
The TOES baselines minimize $t_f$ instead of $J_s$ and are the arbitrary-reference case from Prop. \ref{prop:loading}.
TOER (search-only) uses the reference $\bar{x}_0$ for which the profile holds without qualification. 

The margin over TOES (End @ Base) is the smallest because, at unreliable detection, ending at base is close to return-optimal. 
At $\kappa=10$, TOES (End @ Base) carries an expected return of $0.74$ m against TOER's $0.7$ m, while TOES + Return and Info-Max yield $0.95$ m and $1.01$ m, respectively. 
TOER reaches this terminal behavior without imposing a terminal constraint. 
Since the belief is uniform, Info-Max has no peak to exploit and its detection probability matches the ergodic planners'.
Therefore, the gap between Info-Max and other methods is a result of it ending its search far from base.

Prop. \ref{prop:loading} caps the relative advantage at $1-\bar{d}_{\min}/\bar{d}(\bar{x}_0)$, as set by the target distribution, coverage tolerance, and domain geometry (25.6\% in this experiment).
At $\kappa=10$, TOER achieves $35\%$ of this ceiling. 
The gap to TOER (search-only) isolates the return term alone, with advantage $0.01\%$ at $\kappa=1$ with CI spanning zero and rising to $7.9\%$ at $\kappa=10$.


\subsection{Robustness to Belief Error}
\label{sub:belief}

\begin{figure} [t]
    \centering
    \vspace{14pt}
    \includegraphics[width=\linewidth]{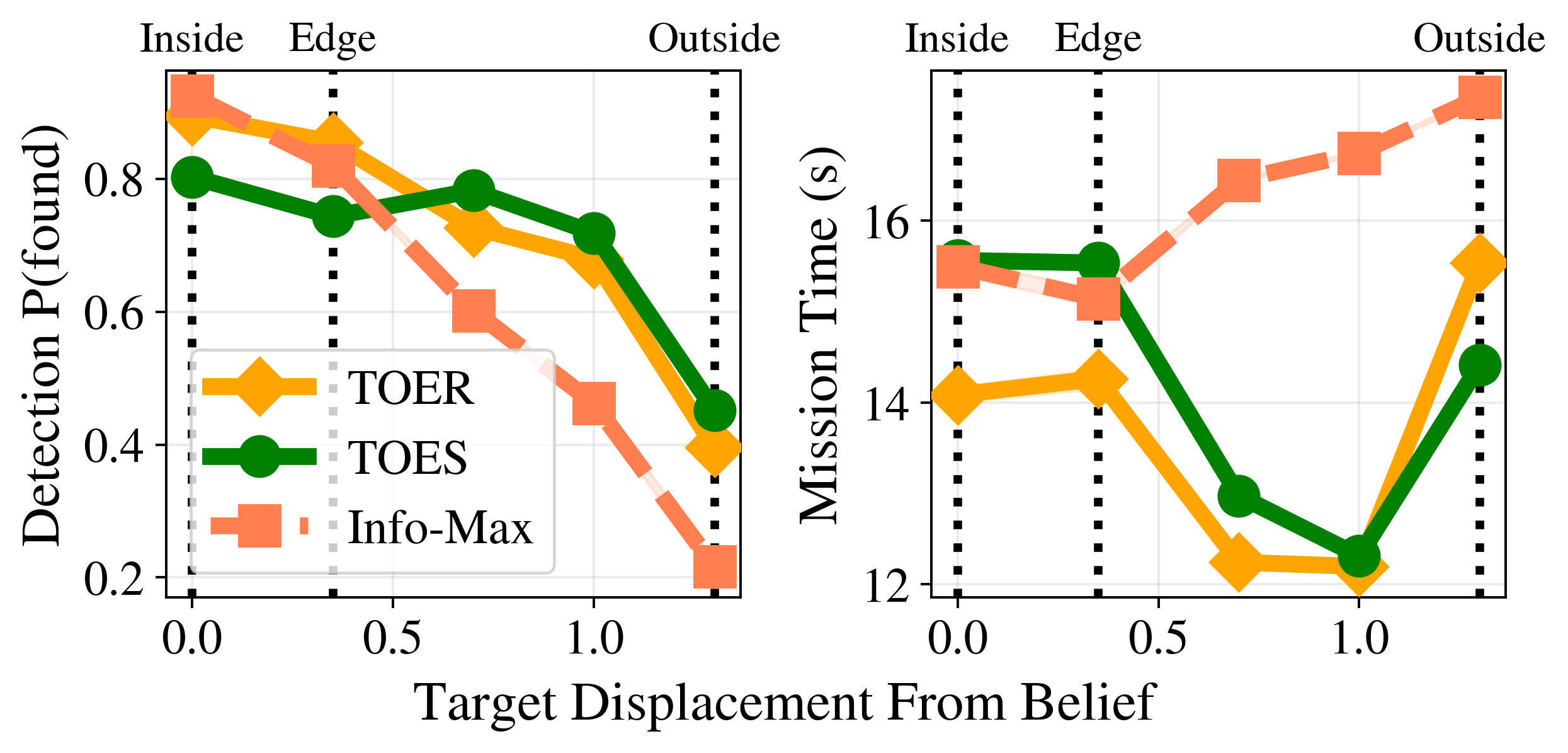}
    \vspace{-22pt}
    \caption{\textbf{Belief Fidelity Impact on Mission.} Search in the room setting from Fig. \ref{fig:e3} as the true target is displaced by $\delta$ from the given belief. (Left) Info-Max's detection collapses once the belief is wrong ($0.93\to0.22$) while the ergodic methods degrade slowly, since covering the belief's support leaves detection mass where the target can still be found. (Right) TOER holds the lowest mission time until the target leaves the belief entirely, where its trade-off between detection speed and terminal position is no longer beneficial.}
    \label{fig:e4}
\end{figure}

We now examine the robustness of each method to prior misspecification to determine each method's reliance on the correctness of the given target distribution (see Fig. \ref{fig:e4}). 

\textbf{Problem Setup.} This experiment uses the same domain as Sec. \ref{sub:loading} with $h_\text{sensor} = 0.023$, a gain of $4.5$, $\kappa=3.3$, and 6 initializations scored on the displaced sample. 
All methods are given the same  2-part Gaussian mixture belief of bandwidth $0.4$ centered on 2 rooms, while the true target is drawn from a Gaussian with $\sigma=0.12$ and displaced by $\delta_\text{target} \in [0, 1.3]$ m.
The coverage kernel is tightened to $h=0.005$ so the ergodic planners do not over-smooth the prior.

When the prior is correctly-specified, Info-Max has the strongest detection probability, which is consistent with its priority of high-value regions. 
As the belief degrades, Info-Max's effectiveness drops from $P(\text{found})=0.93\to0.22$, while the focus of ergodic methods on covering the belief's support allows them to lose detection probability more slowly. 

Mission time follows a similar pattern as detection, but with the added benefit of return-awareness. 
TOER is the fastest method at every displacement up to $\delta_\text{target}=1$ and completes the mission in $12.2$s at $\delta_\text{target}=0.7$ against TOES' $13$s and Info-Max's $16.5$s. 
Beyond this range, the target leaves the belief entirely and no planner can allocate search time from the prior. 
TOES is unaffected by this misplaced allocation because return-blindness costs nothing when the belief fails, while TOER's trade-off between detection speed and terminal position costs $7.8\%$ relative to TOES. 
Over the same range, Info-Max's reliance on the belief's peak costs more than half of its detection probability. 

\subsection{Bound \& Certificate Validation}

\begin{figure} [t]
    \centering
    \vspace{14pt}
    \includegraphics[width=\linewidth]{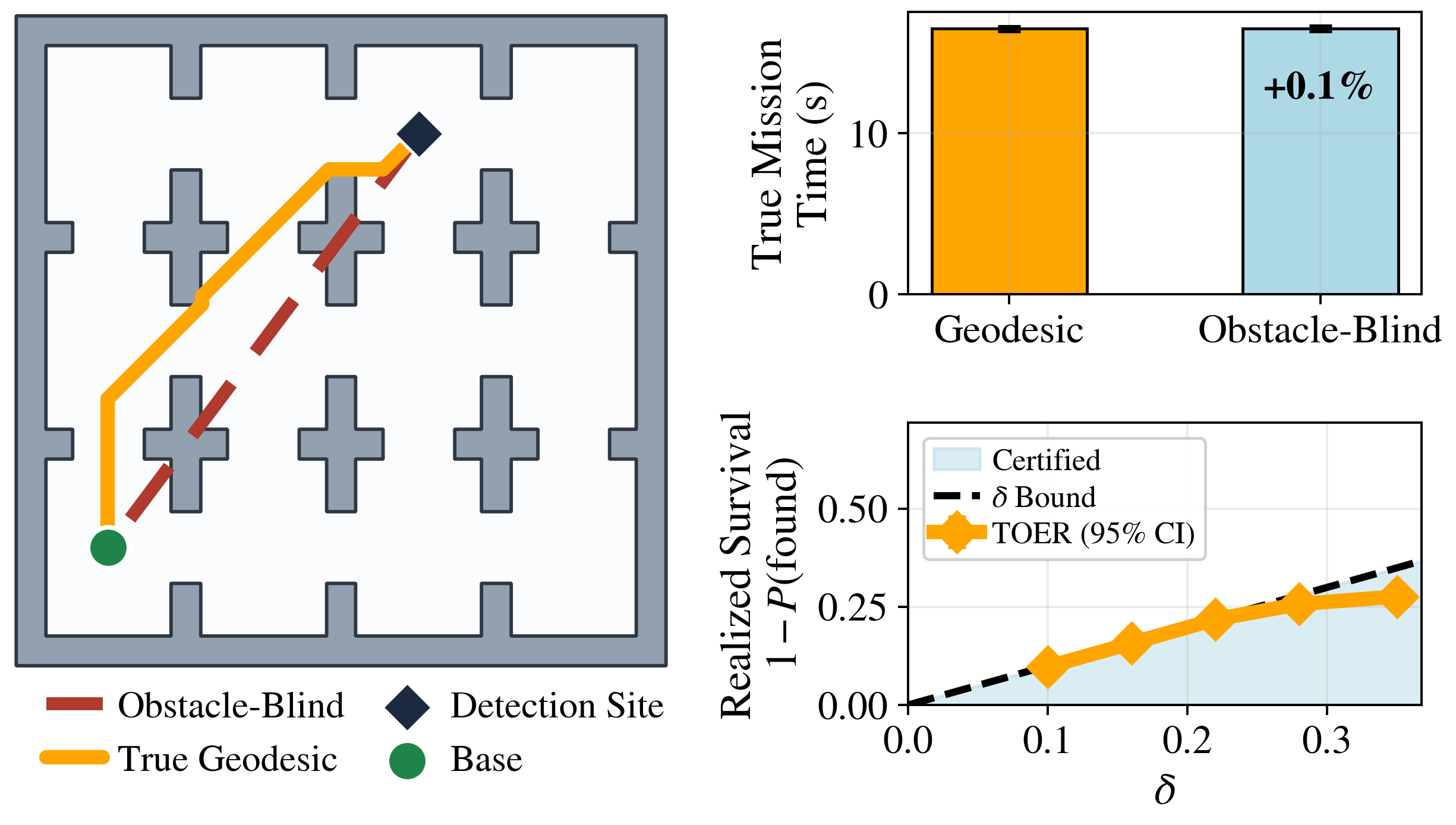}
    \vspace{-22pt}
    \caption{\textbf{Return Bound and Detection Certificate.} Validation of the return bound and detection certificate on the non-convex room grid from Fig. \ref{fig:e3}. (Left) The two return-cost estimates from one detection site. The obstacle-blind bound cuts through walls while the true geodesic routes around them. (Right, Top) Planning against the loose bound costs $0.1\%$ of realized mission time relative to the geodesic bound. (Right, Bottom) With survival $\leq\delta$ imposed, realized survival stays inside the certified region at every $\delta$.}
    \label{fig:e5_guarantees}
\end{figure}

Next, we verify the return bound and certificate validation guarantees numerically. 

\textbf{Problem Setup.} A robot is placed in the same environment as Sec. \ref{sub:loading} with $h_\text{sensor}=0.048$. 
The return bound experiment defines $\kappa=6.6$, computes two return lower bounds inside the objective (obstacle-blind Euclidean distance and a geodesic return), and evaluates both bounds against the true geodesic return. 
The certificate study defines $\kappa=3.33$ and imposes $\Ps(\tf)\le\delta$ as a hard constraint for $\delta\in\{0.35, 0.28, 0.22, 0.16, 0.1\}$. 

As shown in Fig. \ref{fig:e5_guarantees}, the obstacle-blind estimate is at most the geodesic return for $99.3\%$ of sampled states with a mean gap of $0.57$s, which confirms the ordering stated in Prop. \ref{prop:bound}.
The remaining $0.7\%$ of samples violate the ordering by at most $0.014$ s, which is below the $0.085$ s corresponding to one cell diagonal of the geodesic grid. 
These discrepancies are therefore artifacts of the discretization. 
Planning against a looser bound only costs $0.1\%$ of realized mission time with overlapping confidence intervals. 
Therefore, the guarantee holds without computing the geodesics.

In sweeping $\delta$ under the hard survival constraint in Prop. \ref{prop:certificate}, every point and their confidence intervals lie inside the certified region, including $\delta=0.1$, which yields a realized survival of $0.098$. 


\subsection{Computational Scaling}

\begin{figure} [t]
    \centering
    \vspace{14pt}
    \includegraphics[width=\linewidth]{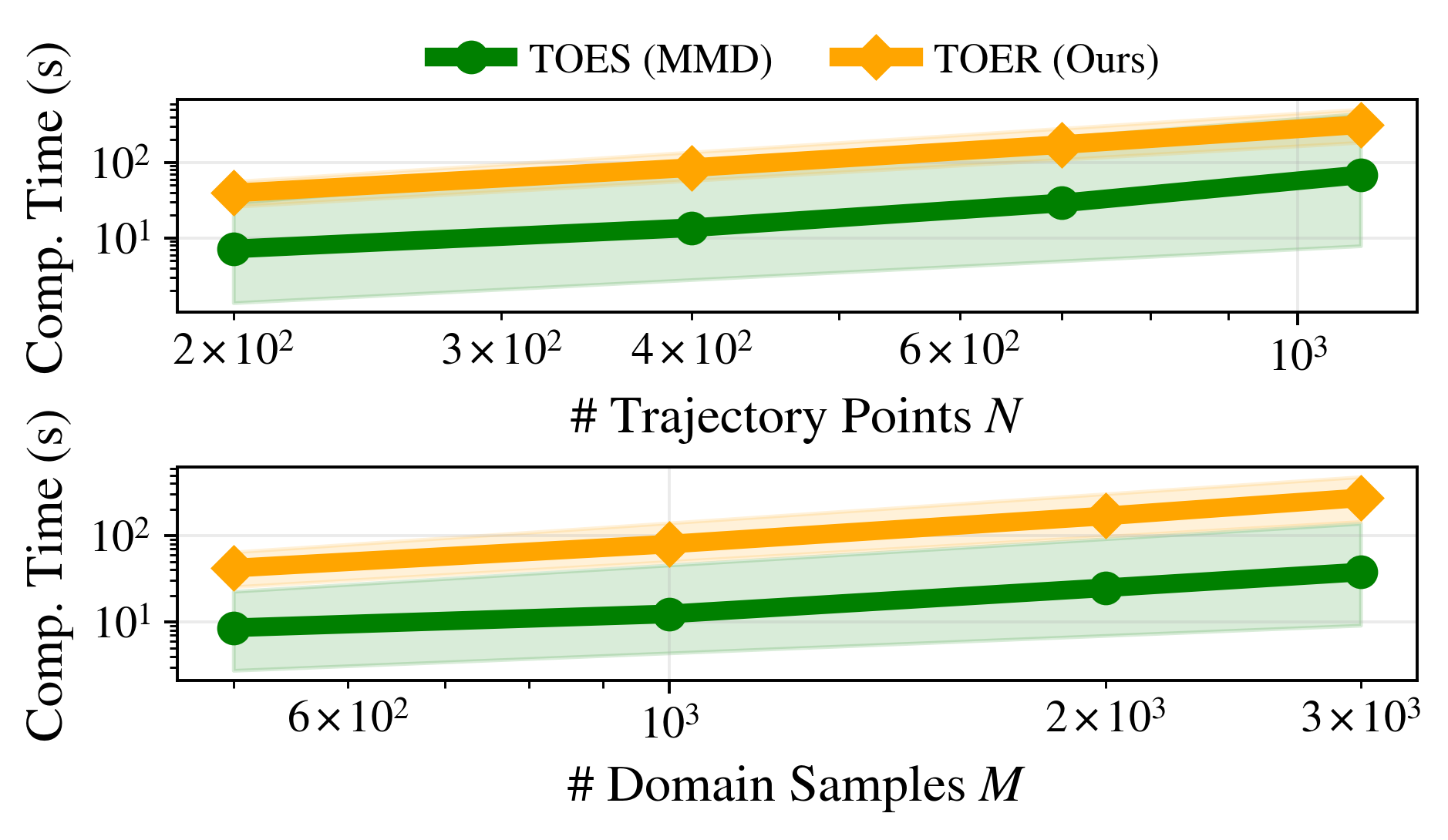}
    \vspace{-24pt}
    \caption{\textbf{Computational Scaling of TOER.} Log-log wall time for a single solve at fixed iterations ($95\%$ CI bands on each slope). TOER and TOES scale identically in trajectory knots $N$, while the Poisson detection term makes TOER grow linearly in domain samples $M$ while TOES is sublinear.}
    \label{fig:e5_scaling}
\end{figure}

In this experiment, we measure what the addition of stochastic detection in the objective costs in computation. 

\textbf{Problem Setup.} We time a single solve at a fixed budget of 3000 solver iterations with sweeps over trajectory length $N\in\{200, 400, 700, 1100\}$ ($M=1500$) and the number of domain samples $M\in\{500, 1000, 2000, 3000\}$ ($N=500$), respectively. 
Experiments are run on a CPU with $\kappa=3.33$.

Fig. \ref{fig:e5_scaling} shows that TOER and TOES scale identically in trajectory length with TOER a near-constant $6\times$ more expensive per-iteration. 
However, because the Poisson survival term is a cumulative sum that must be evaluated at every sample before reduction, TOER grows in computational cost $\approx16\%$ per doubling of $M$ relative to TOES. 
Since both methods are intended to plan offline and $M$ is chosen by the user, this cost does not recur during implementation. 
Solving TOER's search program without the return term costs $125.6$s against TOER's $126.9$s, which verifies that the computational cost is primarily due to the Poisson objective. 

\subsection{Hardware Validation}
Finally, we deploy TOER on hardware to confirm that the formulation transfers unmodified to a physical platform and a 3D search domain (Fig. \ref{fig:abstract}). 
A Franka arm searches a cluttered bin ($\Omega\subset\mathbb{R}^3$) for a target that is fully occluded at the start with detection from an overhead depth camera. 
The plan is solved once before execution, and upon detection of the object, the arm abandons its remaining path, collects the object, and delivers it to the base (white '$\times$') while respecting robot dynamical constraints.


\section{Conclusion}

In this work, we present a formulation of the time-optimal ergodic search problem that plans for both the search and return components of a retrieval mission. 
While existing methods decide where to search but stop when the target is found, we instead minimize the expected time to find and deliver the target, with detection modeled as a Poisson process and ergodicity kept as a constraint. 
Because the return cost enters only the objective, the coverage guarantees of time-optimal ergodic search carry over unchanged.
Results show that our approach reduces mission time relative to baselines and yields mission time reductions up to a fixed ceiling as return becomes more costly. 
A survival constraint additionally lets a user specify a detection confidence and determine the resultant cost in mission time before deployment.
Future work will extend the method to single-shot planning for multiple retrievals under a given capacity constraint.

\bibliographystyle{IEEEtran}
\bibliography{references}

\end{document}